\documentclass[a4paper,twoside]{article}

\usepackage{epsfig}
\usepackage{subcaption}
\usepackage{calc}
\usepackage{amssymb}
\usepackage{amstext}
\usepackage{amsmath}
\usepackage{amsthm}
\usepackage{multicol}
\usepackage{pslatex}
\usepackage{apalike}

\usepackage{cite}
\usepackage{amsmath,amssymb,amsfonts}
\usepackage{algorithmic}
\usepackage{graphicx}
\usepackage{textcomp}
\usepackage{xcolor}
\usepackage{epsfig}
\usepackage{tabu}
\usepackage{multirow} 
\usepackage{adjustbox}
\usepackage{multirow} 
\usepackage{filecontents}
\usepackage{todonotes}
\usepackage{float}
\usepackage{url}
\usepackage{lipsum}
\usepackage{multicol}

\usepackage{pgfplots}

\usepackage{SCITEPRESS}     

\begin{document}

\title{Hallucinating Saliency Maps for Fine-Grained Image Classification for Limited Data Domains} 

 \author{\authorname{Carola Figueroa-Flores\sup{1,2}, Bogdan Raducanu \sup{1} ,  David Berga \sup{1}  and Joost van de Weijer \sup{1}}
 \affiliation{\sup{1}Computer Vision Center\\
Edifici ``O'' - Campus UAB\\
8193 Bellaterra (Barcelona), Spain}
 \affiliation{\sup{2}Department of Computer Science and  Information Technology\\ Universidad del B\'io  B\'io, Chile}
\email{\{cafigueroa, bogdan, dberga, joost\}@cvc.uab.es, cfiguerf@ubiobio.cl}
 }

\keywords{Fine-grained image classification, Saliency detection, Convolutional neural networks
}

\abstract{It has been shown that saliency maps can be used to improve the performance of object recognition systems, especially on datasets that have only limited training data. However, a drawback of such an approach is that it requires a pre-trained saliency network. In the current paper, we propose an approach which does not require explicit saliency maps to improve image classification, but they are learned implicitely, during the training of an end-to-end image classification task. We show that our approach obtains similar results as the case when the saliency maps are provided explicitely.  We validate our method on several datasets for fine-grained classification tasks (Flowers, Birds and Cars), and show that especially for domains with limited data the proposed method significantly improves the results.} 

\onecolumn \maketitle \normalsize \setcounter{footnote}{0} \vfill

\section{\uppercase{Introduction}}
\label{sec:introduction}
Fine-grained image recognition has as objective to recognize  many subcategories of a super-category. Examples of well-known fine-grained datasets are Flowers \cite{nilsback2008automated}, Cars \cite{krause2013} and Birds \cite{WelinderEtal2010}. The challenge of fine-grained image recognition is that the differences between classes are often very subtle, and only the detection of small highly localized features will correctly lead to the recognition of the specific bird or flower species. An additional challenge of fine-grained image recognition is the difficulty of data collection. The labelling of these datasets requires experts and subcategories can be very rare which further complicates the collection of data. Therefore, the ability to train high-quality image classification systems from few data is an important research topic in fine-grained object recognition. 

\begin{figure*}[tb]
\begin{center}
\includegraphics[width=0.9\textwidth]{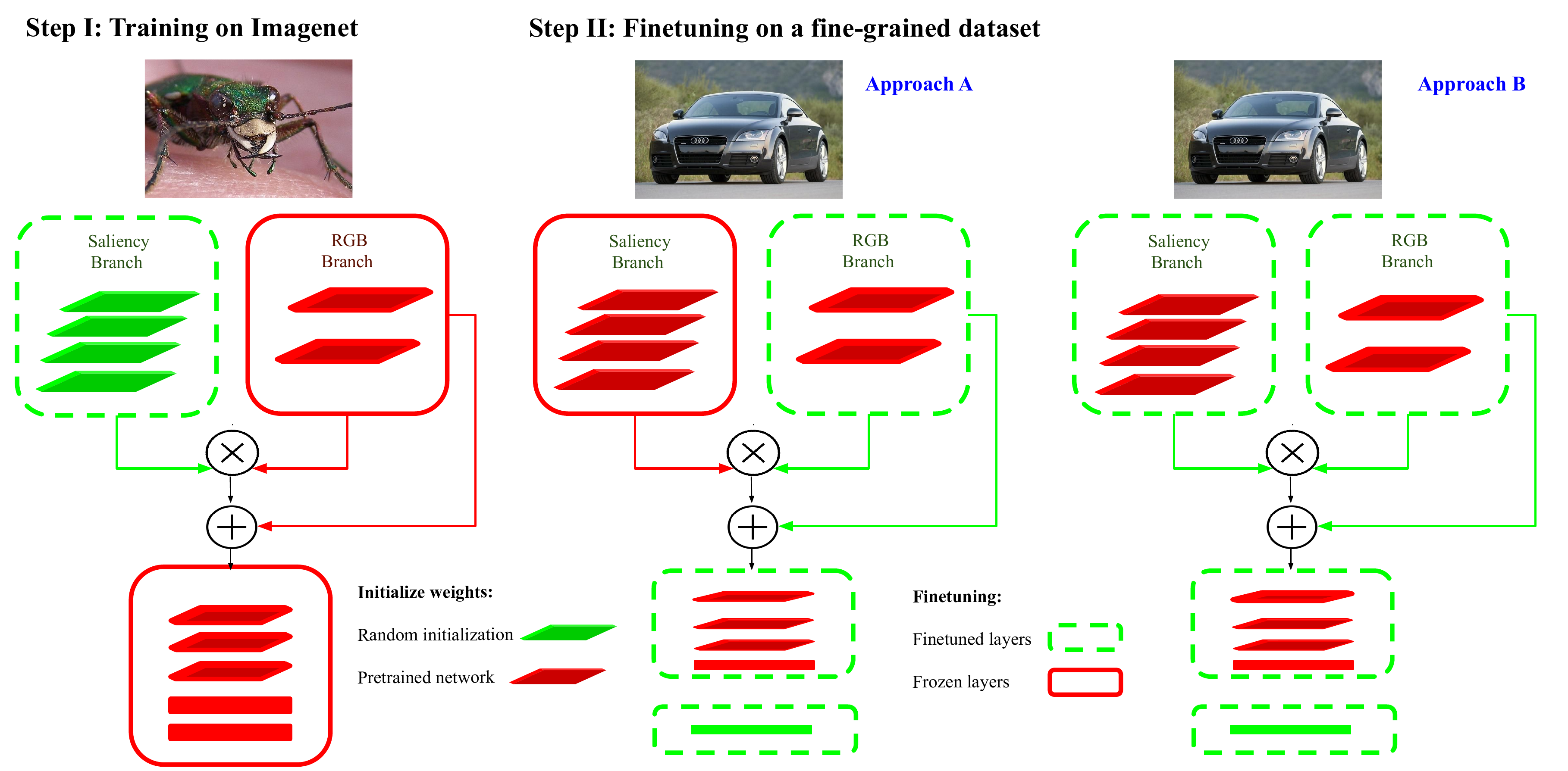}
\end{center}
\caption{Overview of our method. We process an RGB input image through two branches: one branch extracts the RGB features and the other one is used to learn saliency maps. The resulting features are merged via a modulation layer, which continues with a few more convolutional layers and a classification layer. The network is trained in two steps.}
\label{fig:overview}
\end{figure*}

Most of the state-of-the-art general object classification approaches~\cite{wang2017residual,krizhevsky2012imagenet} have difficulties in the fine-grained recognition task, which is more challenging due to the fact that basic-level categories (e.g. different bird species or flowers) share similar shape and visual appearance. Early works have focused on localization and classification of semantic parts using either explicit annotation \cite{zhang2014part,lin2015deeplac,zhang2016spda, ding2019iccv,du2020eccv} or weak labeling \cite{zheng2017iccv,fu2017cvpr,wang2020cvpr}. The main disadvantage of these approaches was that they required two different 'pipelines', for detection and classification, which makes the joint optimization of the two subsystems more complicated. Therefore, more recent approaches are proposing end-to-end strategies with the focus on improving the feature representation from intermediate layers in a CNN through higher order statistics modeling \cite{cai2017iccv,wang2018cvpr}.

One recent approach which obtained good fine-grained recognition results, especially with only few labeled data is proposed in~\cite{Figueroa2019}. The main idea is that a saliency image can be used to modulate the recognition branch of a fine-grained recognition network. We will refer to this technique as saliency-modulated image classification (SMIC). This is especially beneficial when only few labeled data is available. The gradients which are backpropagated are concentrated on the regions which have high saliency. This prevents backpropagation of gradients of uninformative background parts of the image which could lead to overfitting to irrelevant details. A major drawback of this approach is that it requires an explicit saliency algorithm which needs to be trained on a saliency dataset.

In order to overcome the lack of sufficient data for a given modality, a common strategy is to introduce a 'hallucination' mechanism which emulates the effect of genuine data. For instance, in \cite{hoffman2016hallucination}, they use this 'hallucination' strategy for RGB-D object detection. A hallucination network is trained to learn complementary RGB image representation which is taught to
mimic convolutional mid-level features from a depth network. At test time, images are processed jointly through the RGB and hallucination networks, demonstrating an improvement in detection performance. This strategy has been adopted also for the case of few-shot learning \cite{girshick2017hallucinating,wang2018imaginary,zhang2019hallucination}. In this case, the hallucination network has been used to produce additional training sample used to train jointly with the original network (also called a neta-learner).

In this paper, we address the major drawback of SMIC, by implementing a hallucination mechanism in order to remove the requirement for providing saliency images for training obtained using one of the existing algorithms \cite{mit-saliency-benchmark}. In other words,  we show that the explicit saliency branch which requires training on a saliency image dataset, can be replaced with a branch which is trained end-to-end for the task of image classification (for which no saliency dataset is required). We replace the saliency image with the input RGB image (see Figure~\ref{fig:overview}). We then pre-train this network for the task of image classification using a subset from ImageNet validation dataset. During this process, the saliency branch will learn to identify which regions are more discriminative. In a second phase, we initialize the weights of the saliency branch with these pre-trained weights. We then train the system end-to-end on the fine-grained dataset using only the RGB images. Results show that the saliency branch improves fine-grained recognition significantly, especially for domains with few training images.

We briefly summarize below our main contributions:
\begin{itemize}
\item we propose an approach which hallucinates saliency maps that are fused together with the RGB modality via a modulation process,
\item our method does not require any saliency maps for training (like in these works~\cite{murabito2018top,Figueroa2019}) but instead is trained indirectly in an end-to-end fashion by training the network for image classification, 
\item our method improves classification accuracy on three fine-grained datasets, especially for domains with limited data. 

\end{itemize} 

The paper is organized as follows. Section II is devoted to review the related work in fine-grained image classification and saliency estimation. Section III presents our approach. We report our experimental results in Section IV. Finally, Section~V contains our conclusions.

\section{\uppercase{Related work}}

\subsection{Fine-grained image classification}

A first group of approaches on fine-grained recognition operate on a two-stage pipeline: first detecting some object parts and then categorizing the objects using this information. 

The work of \cite{huang2016part} first localizes a set of part keypoints, and then simultaneously processes part and object information to obtain highly descriptive representations. 
Mask-CNN~\cite{Wei2018maskcnn} also aggregates descriptors for parts and objects simultaneously, but using pixel-level masks instead of keypoints. The main drawback of these models is the need of human annotation for the semantic parts in terms of keypoints or bounding boxes. To partially alleviate this tedious task of annotation, \cite{xiao2015cvpr} propose an weakly-supervised approach based on the combination of three types of attention in order to guide the search for object parts in terms of 'what' and 'where'. A further improvement has been reported in \cite{zhang2016cvpr}, where the authors propose and approach free of any object / part annotation. Their method explores a unified framework based on two steps of deep filter response picking. On the other hand, \cite{wang2020cvpr} propose an end-to-end discriminative feature-oriented
Gaussian Mixture Model (DF-GMM) to
learn low-rank feature maps which alleviates the discriminative region diffusion problem in high-level feature maps and thus find better fine-grained details.

A second group of approaches merges these two stages into an end-to-end learning framework which optimize simultaneously both part localization and fine-grained classification. This is achieved by first finding the corresponding parts and then comparing their appearance \cite{wang2018cvpr}. In \cite{xie2017lg}, their framework first performs unsupervised part candidates discovery and global object discovery which are subsequently fed into a two-stream CNN in order to model jointly both the local and global features. In \cite{chen2019cvpr}, they propose an approach based on 'Destruction and Construction Learning' whose purpose is to force the network to understand
ç¡`the semantics of each region. For destruction, a region confusion mechanism (RCM) forces the classification network to learn from discriminative regions. For construction, the region alignment network restores the original region layout by modeling the semantic correlation among regions. A similar idea has been pursued in \cite{du2020eccv}, where they propose a progressive training strategy to encourage the network to learn features at different granularities (using a random jigsaw patch generator) and afterwards fuse them together. Some other works introduce an attention mechanism. For instance, \cite{zheng2017iccv} propose a novel part learning approach by a multi-attention convolutional neural network (MA-CNN) without bounding box/part annotations. MA-CNN jointly learns part proposals (defined as multiple attention areas with strong discrimination ability) and the feature representations on each part. Similar approaches have been in reported in \cite{sun2018eccv,luo2019iccv}. In \cite{ding2019iccv}, they propose a network which learns sparse attention from class peak responses (which usually corresponds to informative object parts) and implements spatial and semantic sampling. Finally, in \cite{ji2020cvpr}, the authors present an attention convolutional binary
neural tree in a weakly-supervised approach. Different root-to-leaf paths in the tree network
focus on different discriminative regions using the attention transformer inserted into the convolutional operations along edges of the tree. The final decision is produced as the summation of the predictions from the leaf nodes. 

In another direction, some end-to-end frameworks aim to enhance the intermediate  representation learning capability of a CNN by encoding higher-order statistics.  For instance in \cite{gao2016cvpr} they capture the second-order information by taking the outer-product over the network output and itself. Other approaches focuses on reducing the high feature dimensionality \cite{kong2017cvpr} or extracting higher order information with kernelized modules \cite{cai2017iccv}. In \cite{wang2018cvpr}, they learn a bank of convolutional filters that capture class-specific discriminative patches without extra part or bounding box annotations. The advantage of this approach is that the network focuses on classification only and avoids the trade-off between recognition and localization.

Regardless, most fine-grained approaches use the object ground-truth bounding box at test time, achieving a significantly lower performance when this information is not available. Moreover, automatically discovering discriminative parts might require large amounts of training images. Our approach is more general, as it only requires image level annotations at training time and could easily generalize to other recognition tasks.

\subsection{Saliency estimation}

Initial efforts in modelling saliency involved multi-scale representations of color, orientation and intensity contrast. These were often biologically inspired such as the well-known work by Itti et al. ~\cite{itti1998model}. 
From that model, a myriad of models were based on handcrafting these features in order to obtain an accurate saliency map \cite{Borji2013c,Bylinskii2015}, either maximizing \cite{Bruce2005} or learning statistics of natural images \cite{Torralba2006,harel2006}. Saliency research was propelled further by the availability of large data sets which enabled the use machine learning algorithms \cite{Borji2018}, mainly pretrained on existing human fixation data. 

The question of whether saliency is important for object recognition and object tracking has been raised in \cite{vasconcelos2010saliency}. Latest methods \cite{Borji2018} take advantage of end-to-end convolutional architectures by finetuning over fixation prediction~\cite{Kummerer2016, Pan_2017_SalGAN2,cornia2018predicting}. But the main goal of these works was to estimate a saliency map, not how saliency could contribute to object recognition. 
In this paper instead, we propose an approach which does not require explicit saliency maps to improve image classification, but they are learned implicitly, during the training of an end-to-end image classification task. We show that our approach obtains similar results as the case when the saliency maps are provided explicitely. 

\section{\uppercase{Proposed Method}}

Several works have shown that having the saliency map of an image can be helpful for object recognition and fine-grained recognition in particular~\cite{murabito2018top,Figueroa2019}. The idea is twofold: the saliency map can help focus the attention on the relevant parts of the image to improve the recognition, and it can help guide the training by focusing the backpropagation to the relevant image regions. In ~\cite{Figueroa2019}, the authors show that saliency-modulated  image  classification (SMIC) is especially efficient for training on datasets with few labeled data. The main drawback of these methods is that they require a trained saliency method. Here we show that this restriction can be removed and that we can hallucinate the saliency image from the RGB image. By training the network for image classification on the imageNet dataset we can obtain the saliency branch without human groundtruth images.

\subsection{Overview of the Method}
The overview of our proposed network architecture is illustrated in Figure~\ref{fig:overview}.
Our network consists of two branches: one to extract the features from an RGB image, and the other one (saliency branch) to generate the saliency map from the same RGB image.
Both branches are combined using a \emph {modulation layer} (represented by the $\times$ symbol) and are then processed by several shared layers of the joint branch which finally ends up with a classification layer. The RGB branch followed by the joint branch resembles a standard image classification network.
The novelty of our architecture is the introduction of the saliency branch, which transforms the generated saliency image into a \emph{modulation image}.
This modulation image is used to modulate the characteristics of the RGB branch, putting more emphasis on those characteristics that are considered important for the fine-grained recognition task.
In the following sections we provide the details of the network architecture, the operation of the modulation layer, and finally, how our saliency map is generated. We explain our model using AlexNet~\cite{krizhevsky2012imagenet} as the base classification network, but the theory could be extended to other convolutional neural network architectures. For instance, in the experimental results section, we also consider the ResNet-152 architecture~\cite{Resnet50}.

\subsection{Hallucination of saliency maps from RGB images}

The function of the visual attention maps is to focus on the location of the characteristics necessary to identify the target classes, ignoring anything else that may be irrelevant to the classification task. Therefore, given an input RGB image, our saliency branch should be able to produce a map of the most salient image locations useful for classification purposes. 

To achieve that, we apply a CNN-based saliency detector consisting of four convolutional layers (based on the AlexNet architecture)\footnote{We vary the number of convolutional layers in the experimental section and found four to be optimal.}. The output from the last convolutional layer, i.e. one with 384 dimensional feature maps with a spatial resolution of 13 × 13 (for a 227 × 227 RGB input image), is further processed using a 1 × 1 convolution and then a function of activation ReLU. This is to calculate the saliency score for each "pixel" in the feature maps of the previous layer, and to produce a single channel map. Finally, to generate the input for the subsequent classification network, the 13 × 13 saliency maps are upsampled to 27 × 27 (which is the default input size of the next classification module) through bilinear interpolation. We justify the size of the output maps by claiming that saliency is a primitive mechanism, used by humans to direct attention to objects of interest, which is evoked by coarse visual stimuli. Therefore, our experiments (see section IV) show that 13 × 13 feature maps can encode the information needed to detect salient areas and drive a classifier with them.

\subsection{Fusion of RGB and Saliency Branches}

Consider an input image $ I (x, y, z) $, where $ z = \{1,2,3\} $ indicate the three color channels of the image. Also consider a saliency map $ s (x, y)$. In Flores et al.~\cite{Figueroa2019}, a network $h\left( I, s\right)$ was trained which performed image classification based on the input image $I$ and the saliency map $s$. Here, we replace the saliency map (which was generated by a saliency algorithm) by a hallucinated saliency map $h\left( I, \mathring s \left( I \right) \right)$. The hallucinated saliency map $\mathring s$ is trained end-to-end and estimated from the same input image $I$ without the need of any ground truth saliency data.

The combination of the hallucinated saliency map $\mathring s$ , which is the output of the saliency branch, and the RGB branch is done with modulation. Consider the output of the $ i ^ {th} $ layer of the network, $ l ^ i $, with dimension $w_i \times h_i \times z_i $. Then we define the modulation
as
\begin{equation}
\hat l^i \left( {x,y,z} \right) = l^i \left( {x,y,z} \right) \cdot \mathring s\left( {x,y}\right),  \label{eq:forward}
\end{equation}
resulting in the saliency-modulated layer $\hat l^i$. Note that a single hallucinated saliency map is used to modulate all $i$ feature maps of $\hat l$.  
 
In addition to the formula in Eq.~\eqref{eq:forward} we also introduce a skip connection from the RGB branch to the beginning of the joint branch, defined as
\begin{equation}
\hat l^i \left( {x,y,z} \right) = l^i \left( {x,y,z} \right) \cdot \left( \mathring s\left( {x,y} \right) + 1 \right)\label{eq:forward2}.
\end{equation}
This skip connection is depicted in Figure ~\ref{fig:overview} (+ symbol). 
It prevents the modulation layer from completely ignoring the features from the RGB branch. This is inspired by this work~\cite{Figueroa2019} that found this approach beneficial when using attention for network compression. 

We train our architecture in an end-to-end manner. The backpropagated gradient for the modulation layer into the image classification branch is equal defined as:
\begin{equation}
\frac{{\partial L}}{{\partial l^i }} = \frac{{\partial L}}{{\partial \hat l^i }} \cdot \left( \mathring s\left( x,y \right) +1 \right),
\end{equation}
where $L$ is the loss function of the network. We can see that the saliency map modulates both the forward pass (see Eq.~\eqref{eq:forward2}) as well as the backward pass in the same manner; in both cases putting more weight on the features that are on locations with high saliency, and putting less weight on the irrelevant features. We show in the experiments that this helps the network train more efficiently, also on datasets with only few labeled samples. The modulation prevents the network from overfitting to the background. 

\begin{figure}[tb]
    \centering \;\;
\begin{tikzpicture}[scale=0.65,line width=1pt]
           \begin{axis}[
               bar width=24pt,
               symbolic x coords={Conv-1, Conv-2, Conv-3, Conv-4, Baseline},
               xtick=data,
               x tick label style={rotate=45,anchor=east},
             ]
               \addplot[ybar,fill=cyan] coordinates {
                   (Conv-1,   87.8)
                   (Conv-2,  89.1)
                   (Conv-3,   91.2)
                   (Conv-4,   92.4)
                   (Baseline,   87.8)
               };
           \end{axis}
           \label{chart:exp1}
\end{tikzpicture}
    \caption{Graph shows the classification accuracy on Flowers for various number of layers in the saliency branch. Best results are obtained with four convolutional layers. Baseline refers to the method without saliency branch.}
    \label{fig:optimal_layer}
\end{figure}
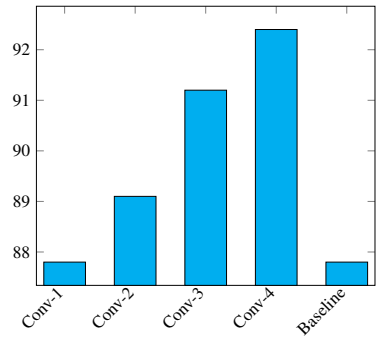

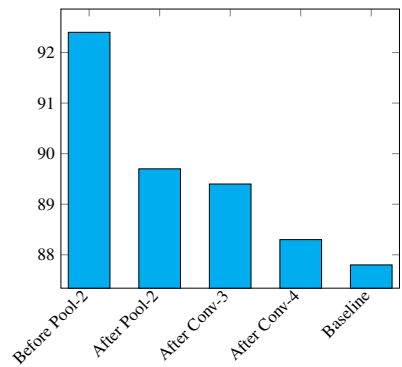
\begin{figure}[tb]
    \centering
    
\begin{tikzpicture}[scale=0.65,line width=1pt]
           \begin{axis}[
               bar width=24pt,
               symbolic x coords={Before Pool-2, After Pool-2, After Conv-3, After Conv-4, Baseline},
               xtick=data,
               x tick label style={rotate=45,anchor=east},
             ]
               \addplot[ybar,fill=cyan] coordinates {
                   (Before Pool-2,   92.4)
                   (After Pool-2,  89.7)
                   (After Conv-3,   89.4)
                   (After Conv-4,   88.3)
                   (Baseline,   87.8)

               };
               
           \end{axis}
           \label{chart:exp2}
\end{tikzpicture}
    \caption{Graph shows the classification accuracy on Flowers. Various points for fusing the saliency and RGB branch are evaluated. Best results are obtained when fusion is placed before the pool-2 layer. Baseline refers to the method without saliency branch.}
    \label{fig:optimal_fusion}
\end{figure}

\begin{table*}[tb]
\centering
\resizebox{1.0\textwidth}{!}{%
        \begin{tabular}{l|l|ccccccccccc}
        \hline

            \hline
            \parbox[t]{2mm}{\multirow{6}{*}{\rotatebox[origin=c]{90}{Flowers}}} 
             
            & \#train images  & 1 & 2 & 3 & 5 & 10 & 15 & 20 & 25 & 30 & $K$ & $\textbf{AVG}$\\ \hline

           &  Baseline-RGB & 31.8 & 45.8 & 53.1 & 63.6 & 72.4 & 76.9 & 81.2 & 85.1 & 87.2 & 87.8 & 68.3\\ 

           &  Baseline-RGB + scratch SAL & 34.3 &48.9 & 54.3  & 65.9  & 73.1 & 77.4  &82.3  &85.9  &88.9  &89.1 & 70.0\\ 
            
           & SMIC~\cite{Figueroa2019}$^*$  & \textbf{37.6} & \textbf{51.9}  &57.1  &68.5  &75.2  &\textbf{79.7}  &\textbf{84.9}  & 88.2 &91.2  &92.3 & \textbf{72.7} \\ 
            
             & Approach A & 36.9 &51.3  &56.9  &67.8  &74.9  &78.4  &82.9  &88.1  &90.9  &92.0 &72.0 \\
            
            & Approach B  & 37.3  &51.7  &\textbf{57.2}  &\textbf{68.7}  & \textbf{75.6}  & 78.7 &83.8  &\textbf{88.4}  &\textbf{91.7}  &\textbf{92.5} &72.6 \\ \hline

           \hline 
           \parbox[t]{2mm}{\multirow{6}{*}{\rotatebox[origin=c]{90}{Cars}}}
           & Baseline-RGB & 4.1 & 7.8  & 11.7  &17.3  &25.5  &31.1  &38.5  &42.2  &47.2  &60.0  & 28.5\\ 

          & Baseline-RGB + scratch SAL &5.9  &10.7 & 14.4  &19.1   &27.4  &32.9   &38.5  & 44.0 &48.7  &61.5 &30.3 \\
            
           & SMIC~\cite{Figueroa2019}$^*$  &9.3  & 14.0  &18.0  &22.8  &\textbf{30.0}  &34.7  &\textbf{40.4}  &\textbf{46.0}  &\textbf{50.0}  &61.4  & 32.7\\ 
            
          &  Approach A & 9.3 &14.3  &17.4  &22.3  &28.4  &35.3  &39.7  &45.7  & 50.1 &61.9  & 32.4\\
            
          &  Approach B & \textbf{9.8} &\textbf{15.1}  &\textbf{18.4}  &\textbf{22.9}  &28.8  &\textbf{35.1}  &39.9  &45.8  &49.7  &\textbf{62.9}  &\textbf{32.8} \\ \hline

           \hline 
           \parbox[t]{2mm}{\multirow{6}{*}{\rotatebox[origin=c]{90}{Birds}}}
          &  Baseline-RGB &  9.1& 13.6  &19.4  &27.7  & 37.8  & 44.3 & 48.0  &50.0  & 54.2 & 57.0  & 34.8 \\ 

          &  Baseline-RGB + scratch SAL & 10.4 &14.9 &20.3   &28.3   &38.6  &43.9   &46.9  &48.4 & 50.7 &55.7 &35.8 \\ 
           
          & SMIC~\cite{Figueroa2019}$^*$ & \textbf{13.1} & \textbf{18.9}  &22.2  &\textbf{30.2}  &38.7  &44.3  &48.0  &\textbf{50.0}  &\textbf{54.2}  &57.0  & 37.7\\
           
          & Approach A & 11.8 & 18.3 &22.1  &29.3  &39.1  &\textbf{44.4}  &47.8  &49.7  &53.1  &56.5  &37.2 \\
            
          &   Approach B  & 12.9  &18.7  & \textbf{22.7} & 29.7 & \textbf{39.4}  &44.1  &\textbf{48.2}  &49.9  &53.9  &\textbf{57.7}  & \textbf{37.7} \\ \hline

        \end{tabular}

}        
        \caption{Classification accuracy for Flowers, Cars, and Birds dataset (results are the average over three runs), using \textbf{AlexNet} as base network. Results are provided for varying number of training images, from 1 until 30; $K$ refers to using the number of training images used in the official dataset split. The rightmost column shows the average. The $^*$ indicates that the method requires an explicit saliency method. Our method (Approach B) obtains similar results as SMIC but without the need of a pretrained saliency network trained on a saliency dataset.}
        \label{tab:saliencyFlower102}
\end{table*}

\begin{table*}[tb]
\centering
\resizebox{1.0\textwidth}{!}{%
        \begin{tabular}{l|l|ccccccccccc}
        \hline

            \hline
            \parbox[t]{2mm}{\multirow{6}{*}{\rotatebox[origin=c]{90}{Flowers}}} 
             
            & \#train images  & 1 & 2 & 3 & 5 & 10 & 15 & 20 & 25 & 30 & $K$ & $\textbf{AVG}$\\ \hline

           &  Baseline-RGB & 39.0 & 60.1 & 68.0 & 82.5 & 89.0 & 92.0 & 92.1  & 93.3  & 94.2 & 95.4 & 80.3\\ 

           &  Baseline-RGB + scratch SAL & 40.1 & 63.8  & 69.7 & 83.9  & 89.7  & 91.9  & 92.9  & 93.8  & 95.1  & 97.1  & 81.8 \\ 
            
           & SMIC~\cite{Figueroa2019}$^*$  & 42.6 & 64.2  & 70.9  & \textbf{85.5}  & \textbf{90.9}  & \textbf{92.7}  & \textbf{94.0}  & \textbf{95.0} & \textbf{97.0}  & 97.8  & \textbf{83.1} \\ 
            
             & Approach A & 42.4 & \textbf{64.5}  & 70.7  & 85.2  & 90.3  & 92.4  & 93.3  & 94.3 & 96.5  & 97.9  & 82.8 \\
            
            & Approach B  & \textbf{42.7} & \textbf{64.5} & \textbf{71.0}  & 85.1  & 90.4  & 92.5  & 93.1  & 94.7  & 96.8  & \textbf{98.1}  & 82.9\\ \hline

           \hline 
           \parbox[t]{2mm}{\multirow{6}{*}{\rotatebox[origin=c]{90}{Cars}}}
           & Baseline-RGB & 30.9 & 45.8 & 53.1  & 62.7  & 70.9  & 73.9  & 79.9  & 88.7  & 89.2  & 90.7  & 68.6 \\ 

          & Baseline-RGB + scratch SAL & 33.8 & 46.1 & 54.8 & 63.8  & 71.7  & 74.9 & 80.9  & 88.1  & 89.1  & 91.0  & 69.4 \\
            
           & SMIC~\cite{Figueroa2019}$^*$  & \textbf{34.7} & \textbf{47.9}  & 55.2  & \textbf{64.9}  & \textbf{72.1}  & \textbf{75.8}  & \textbf{82.1}  & \textbf{90.0}  & \textbf{91.1} & \textbf{92.4}  & \textbf{70.6}\\ 
            
          &  Approach A & 34.1 & 47.0  & 56.3  & 64.7 & 71.9 & 75.3 & 81.7 & 89.0 & 90.8  & 91.7  & 70.2\\
            
          &  Approach B & 34.0 & 47.5  & \textbf{55.4}  & 64.7  & 71.8  & 75.5 & 81.9 & 89.3 & 91.0 & 92.1 & 70.3 \\ \hline

           \hline 
           \parbox[t]{2mm}{\multirow{6}{*}{\rotatebox[origin=c]{90}{Birds}}}
          &  Baseline-RGB & 24.9 & 35.3  & 44.1 & 53.3 & 63.8 & 71.8 & 75.7 & 79.3 & 82.9 & 83.7 & 61.5\\ 

          &  Baseline-RGB + scratch SAL & 26.3 & 36.1 & 45.2 & 53.9 & 64.3 & 72.1  & 76.3 & 79.9 & 83.1 & 83.4 & 62.1\\ 
           
          & SMIC~\cite{Figueroa2019}$^*$ & \textbf{28.1} & \textbf{37.9} & \textbf{46.8} & \textbf{55.2} & 65.3 & \textbf{73.1} & 77.0 & \textbf{82.9} & \textbf{84.4} & \textbf{86.1} & \textbf{63.7}\\
           
          & Approach A & 26.9 & 36.9 & 46.1 & 54.2 & 64.9 & 72.8 & \textbf{77.1} & 81.4 & 83.4  & 84.8 & 62.9\\
            
          &   Approach B  & 27.1 & 37.0 & 46.2 & 54.9 & \textbf{65.4} & 72.8 & \textbf{77.1} & 81.3 & 83.8 & 85.1 & 63.1 \\ \hline

        \end{tabular}

}        
        \caption{Classification accuracy for Flowers, Cars, and Birds dataset (results are the average over three runs), using \textbf{ResNet152} as base network. Results are provided for varying number of training images, from 1 until 30; $K$ refers to using the number of training images used in the official dataset split. The rightmost column shows the average. The $^*$ indicates that the method requires an explicit saliency method. Our method (Approach B) obtains similar results as SMIC but without the need of a pretrained saliency network trained on a saliency dataset.}
        \label{tab:saliencyResNet}
\end{table*}

\subsection{Training on Imagenet and fine-tuning on a target dataset}
As can be seen in Figure ~\ref{fig:overview}, the training of our approach is divided into two steps: first, training on Imagenet and second, fine-tuning on a target dataset.

\noindent
\textbf{Step 1: Training of saliency branch on Imagenet.}

As explained above, the aim of the saliency branch is to hallucinate (generate) a saliency map directly from an RGB input image. 
This network is constructed by initializing the RGB branch with pretrained weights from Imagenet. The  weights of the saliency branch are initialized randomly using the Xavier method (see Figure~\ref{fig:overview}, left image). The network is then trained selectively, using the ImageNet validation set: we allow to train only the layers corresponding to the saliency branch (depicted by the surrounding dotted line) and freeze all the remaining layers (depicted through the continuous line boxes).

\noindent
\textbf{Step 2: Fine-tuning on a target dataset.}
In this step, we initialize the RGB branch with the weights pre-trained from Imagenet and the saliency branch with the corresponding pre-trained weights from Step 1. The weights of the top classification layer are initialized randomly, using the Xavier method. 
Then, this network is then further fine-tuned on a target dataset, selectively. We distinguish two cases:
\begin{itemize}
\item{Approach A:} 
We freeze the layers of the saliency branch and we allow all the other layers layers in the network to be trained. This process is depicted by the continuous line surrounding the saliency branch and the dotted line for the rest (see the Figure~\ref{fig:overview}, middle image).
\item{Approach B:}
We allow all layers to be trained. Since we consider training on datasets with only few labels this could results in overfitting, since it requires all the weights of the saliency branch to be learned  (see the Figure~\ref{fig:overview}, right image) . 
\end{itemize}
In the experiments we evaluate both approaches to training the network.

\section{\uppercase{Experiments}}
\subsection{Experimental Setup}

\noindent
\textbf{Datasets.} To evaluate our approach, we used three standard datasets used for fine-grained image  classification:
\begin{itemize}
\item  \textit{Flowers}: Oxford Flower 102 dataset~\cite{nilsback2008automated} has 8.189 images divided in 102 classes.

\item \textit{Birds}: CUB200 has 11.788 images of 200 different bird species~\cite{WelinderEtal2010}.

\item \textit{Cars}: the CARS-196 dataset in~\cite{krause2013} contains 16,185 images of 196 car classes. 

\end{itemize}

\noindent
\textbf{Network architectures.} We evaluate our approach using two network architectures: Alexnet~\cite{krizhevsky2012imagenet} and Resnet-152~\cite{Resnet50}. In both cases, the weights were pretrained on Imagenet and then finetuned on each of the datasets mentioned above. The networks were trained for 70 epochs with a learning rate of 0.0001 and a weight decay of 0.005. The top classification layer was initialized from scratch using Xavier method~\cite{glorot2010understanding}.  

\noindent
\textbf{Evaluation protocol.} To validate our approach, we follow the same protocol as in \cite{Figueroa2019}. For the image classification task, we train each model with subsets of $k$ training images for $k\in\{1,2,3,5,10,15,20,25,30,K\}$, where $k$ is the total number of training images for the class. We keep 5 images per class for validation and 5 images per class for test. We report the performance in terms of accuracy, i.e. percentage of correctly classified samples. We show the results as an average over three runs.

\subsection{Fine-grained Image Classification Results}

\textbf{Optimal depth and fusion saliency branch:} First we evaluate the saliency branch with a varying number of convolutional layers. The results are presented in Figure~\ref{fig:optimal_layer}. We found that four convolutional layers led to a significant increase in performance. In addition, we look at the best RGB branch layer to perform the fusion of the saliency branch and the RGB branch. The results are presented in Figure~\ref{fig:optimal_fusion}. It is found to be optimal to fuse the two branches before the Pool-2 layer for AlexNet\footnote{In a similar study, we found that for Resnet-152 the optimal fusion is after the forth residual block.}. Based on these experiments, we use four convolutional layers in the saliency branch and fuse before the second pool layer for the remainder of the experiments and for all datasets.

\textbf{Evaluation on scarce data domain:} As described in section III, we consider two alternative ways to train the saliency branch on the target dataset: keeping the saliency branch fixed (Approach A) or allowing it to finetune (Approach B). In this section, we compare these two approaches with respect to the Baseline-RGB and Baseline-RGB + scratch SAL (where Saliency branch is initialized from scratch without pretraining on Imagenet). In addition, we compare to the SMIC method of Flores et al.~\cite{Figueroa2019} who also reports results for small training datasets. We do not compare to other fine-grained methods here, because they do not report results when only considering few labeled images.  The experiments are performed on \textit{Flowers}, 
 \textit{Cars}  and \textit{Birds} datasets and can be seen in Table~\ref{tab:saliencyFlower102}. The average improvement of accuracy of our \emph{Approach} \emph{A} and \emph{B} with respect the \emph{Baseline-RGB} is 3.7\% and 4.3\%, respectively for the \emph{Flowers} dataset; 3.9\% and 4.3\%, respectively for the \emph{Cars} dataset; and 2.4\% and 2.9\%, respectively for the \emph{Birds} dataset. Our \textit{Approach B}  is especially advantageous when we compare it with the SMIC approach, where an additional algorithm is needed to generate the salience map. It is therefore advantageous to also finetune the saliency branch on the target data even when we only have a few labeled images per class. 
 
In Table~\ref{tab:saliencyResNet}, we show the same results but now for ResNet152. One can see that the results improve significantly, especially for \textit{Cars} results improve a lot. The same general conclusions can be drawn : \textit{Approach B} obtains better results than \textit{Approach A} and the method obtains similar results as SMIC but without the need of a pretrained saliency network.

\noindent\textbf{Qualitative results:}  Table~\ref{table:acierto} shows some qualitative results for the case when the pretrained version of our approach predicts the correct label, meanwhile the Baseline-RGB fails. Alternatively, in Table~\ref{table:error} depicts the opposite case: the Baseline-RGB predicts the correct label of the test images, meanwhile the pretrained version of our approach fails. In both cases, the saliency images have been generated using our Approach B. A possible explanation for the failures in this latter case could be that the saliency images are not able to capture the relevant region of the image for fine-grained discrimination. Thus, the salience-modulated layer focuses on the wrong features for the task.
\begin{table}[tb]
\scriptsize
\centering
\caption{Some success examples on Flowers: when the prediction done by Baseline-RGB fails to infer the right label for some test images, but the prediction by our approach is correct. Example image contains image of the wrongly predicted class.}. 
\begin{tabular}{ |ccc| } 
\hline
Input Image & Our Saliency & Example Image \\
\includegraphics[width=0.7in,height=0.5in]{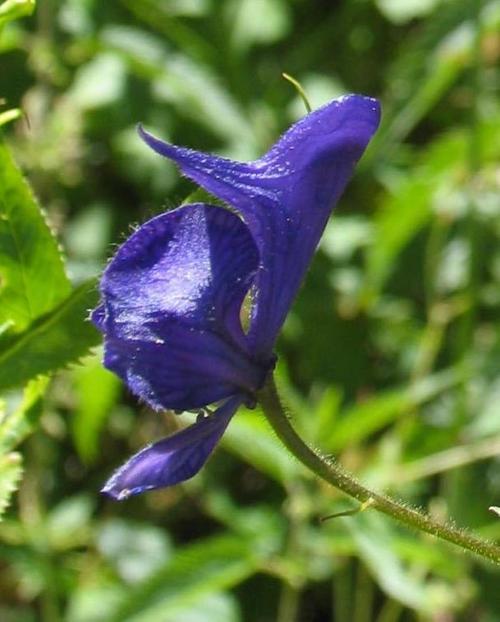} & \includegraphics[width=0.7in,height=0.5in]{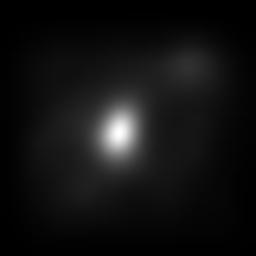} & \includegraphics[width=0.7in,height=0.5in]{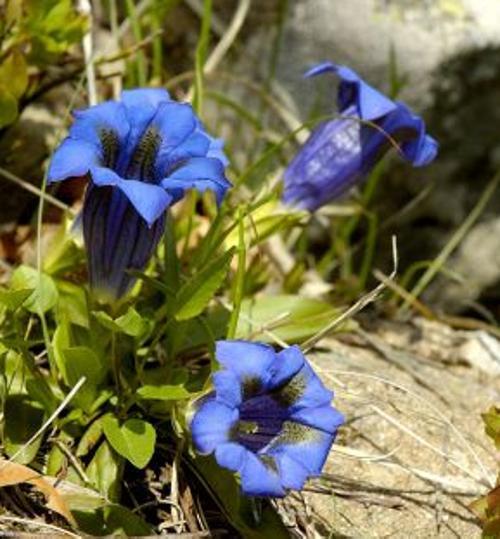} \\
 \multicolumn{3}{ |l| }{Predicted (Baseline-RGB): StemlessGentian } \\ 
  \multicolumn{3}{ |l|} { \textbf{Predicted (Our Approach B): Moonkshood} } \\
   \multicolumn{3}{ |l| }{Ground Truth: Moonkshood } \\ 
  
\includegraphics[width=0.7in,height=0.5in]{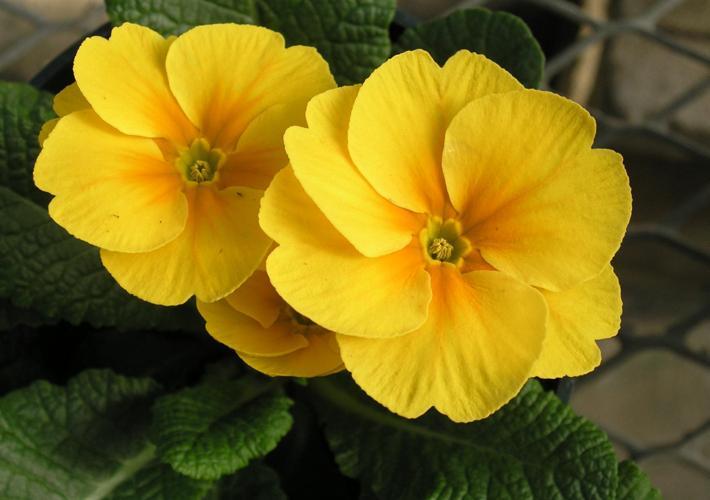} & \includegraphics[width=0.7in,height=0.5in]{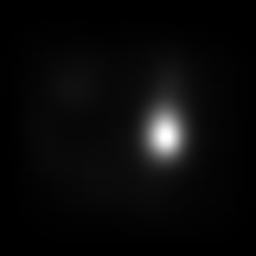} & \includegraphics[width=0.7in,height=0.5in]{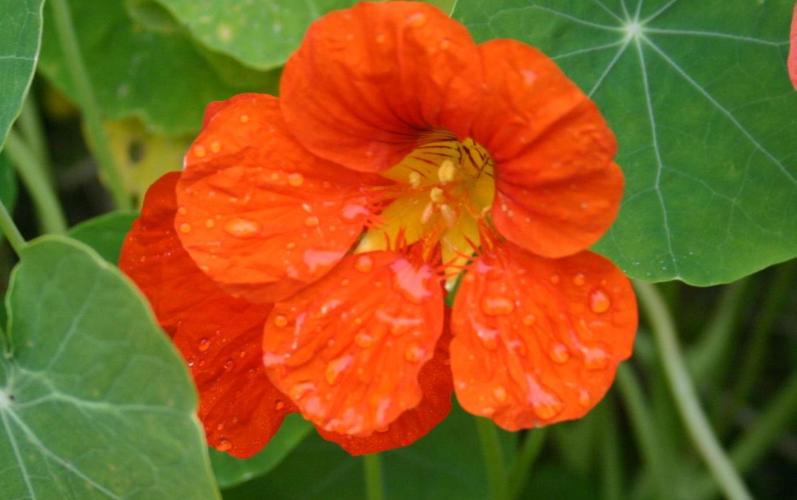} \\
 \multicolumn{3}{ |l| }{Predicted (Baseline-RGB): Watercress } \\ 
  \multicolumn{3}{ |l|}{\textbf{Predicted (Our Approach B): Primula }} \\
   \multicolumn{3}{ |l| }{Ground Truth: Primula } \\ 

\includegraphics[width=0.7in,height=0.5in]{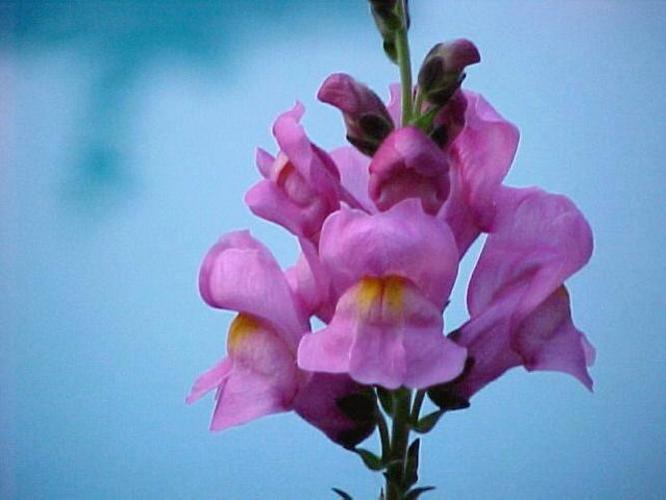} & \includegraphics[width=0.7in,height=0.5in]{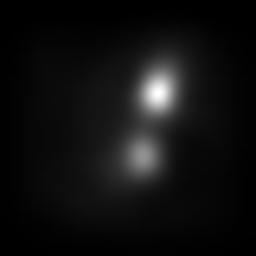} & \includegraphics[width=0.7in,height=0.5in]{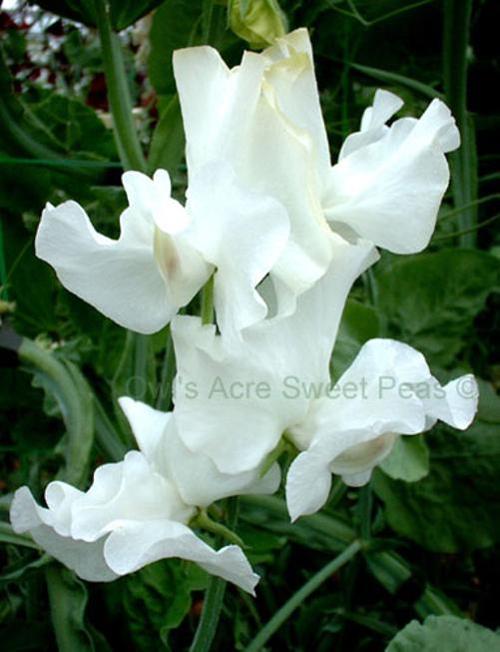} \\
 \multicolumn{3}{ |l| }{Predicted (Baseline-RGB): Sweet Pea } \\ 
  \multicolumn{3}{ |l| }{\textbf{Predicted (Our Approach B): Snap dragon} } \\
   \multicolumn{3}{ |l| }{Ground Truth: Snap dragon } \\  \hline
\end{tabular}
\label{table:acierto}
\end{table}

\section*{\uppercase{Conclusions}}
In this work, we proposed a method to improve fine-graned image classification by means of saliency maps. Our method does not require explicit saliency maps, but they are learned implicitely during the training of an end-to-end deep convolutional network. We validated our method on several datasets for fine-grained classification tasks (Flowers, Birds and Cars). We showed that our approach obtains similar results as the SMIC method~\cite{Figueroa2019} which required explicit saliency maps. We showed that combining RGB data with saliency maps represents a significant advantage for object recognition, especially for the case when training data is limited. 

\begin{table}[tb]
\scriptsize
\centering
\caption{Some failure examples on Flowers: when the prediction done by our method fails to infer the right label for some test images, but the prediction by Baseline-RGB is correct. Example image contains image of the wrongly predicted class.}. 
\begin{tabular}{ |ccc| } 
\hline
Input Image & Our Saliency & Example Image \\
\includegraphics[width=0.7in,height=0.5in]{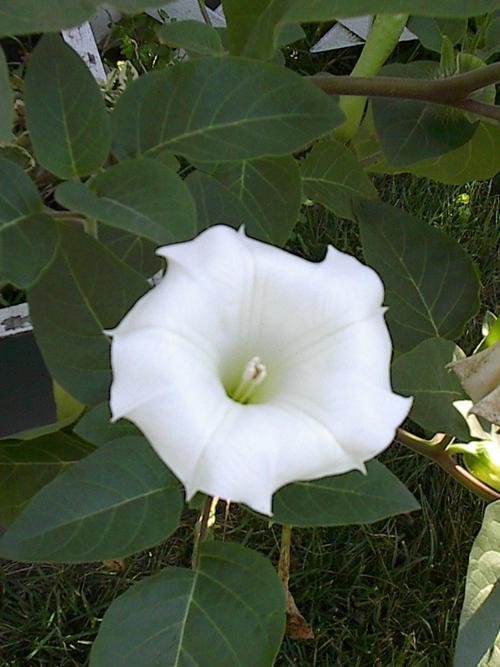} & \includegraphics[width=0.7in,height=0.5in]{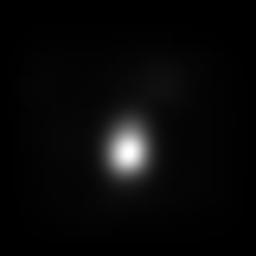} & \includegraphics[width=0.7in,height=0.5in]{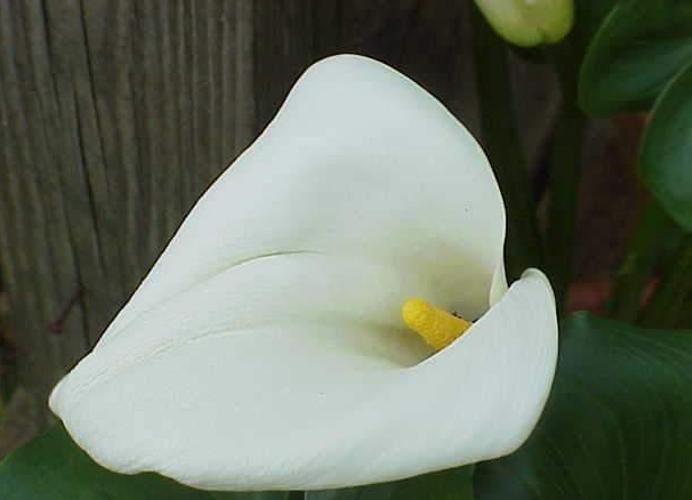} \\
 \multicolumn{3}{ |l| }{\textbf{Predicted (Baseline-RGB): Thorn Apple} } \\ 
  \multicolumn{3}{ |l| }{Predicted (Our Approach B): Arum Lily } \\
   \multicolumn{3}{ |l| }{Ground Truth: Thorn Apple } \\ 
  
\includegraphics[width=0.7in,height=0.5in]{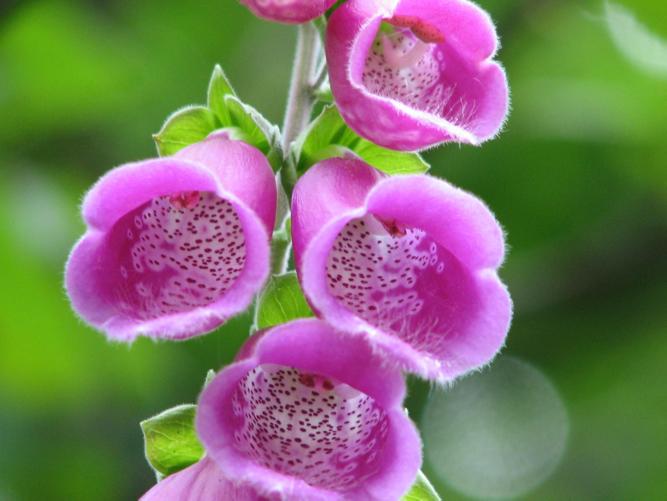} & \includegraphics[width=0.7in,height=0.5in]{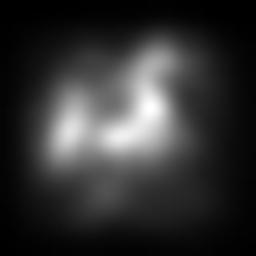} & \includegraphics[width=0.7in,height=0.5in]{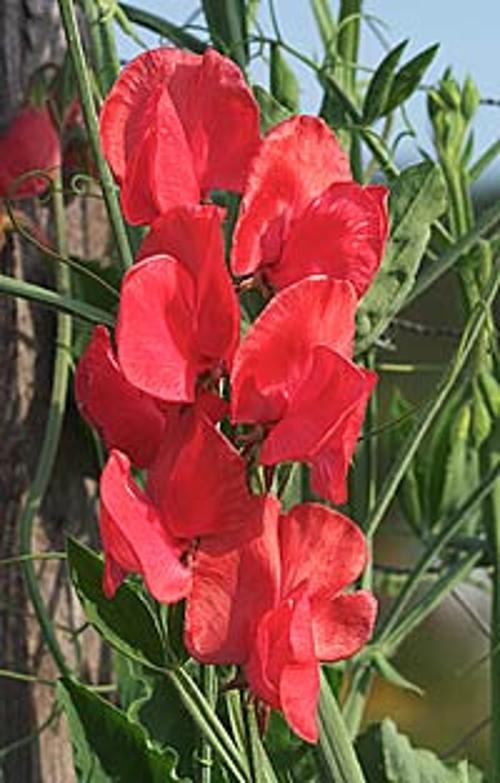} \\
 \multicolumn{3}{ |l| }{\textbf{Predicted (Baseline-RGB): Foxglove }} \\ 
  \multicolumn{3}{ |l| }{Predicted (Our Approach B): Sweet Pea } \\
   \multicolumn{3}{ |l| }{Ground Truth: Foxglove} \\ \hline
\end{tabular}
\label{table:error}
\end{table}

\section*{\uppercase{Acknowledgements}}
The authors acknowledge the Spanish  project  PID2019-104174GB-I00 (MINECO). and the CERCA Programme of Generalitat de Catalunya. Carola Figueroa is supported by a Ph.D. scholarship from CONICYT (now ANID), Chile.
\textit{$\backslash$section*\{ACKNOWLEDGEMENTS\}}

\bibliographystyle{apalike}
{\small
\bibliography{ref}}

\end{document}